\documentclass{article}
\PassOptionsToPackage{numbers,sort&compress}{natbib} % force numeric Natbib
\usepackage[preprint]{paper}
\usepackage[utf8]{inputenc} % allow utf-8 input
\usepackage[T1]{fontenc}    % use 8-bit T1 fonts
\usepackage{hyperref}       % hyperlinks
\usepackage{url}            % simple URL typesetting
\usepackage{booktabs}       % professional-quality tables
\usepackage{amsfonts}       % blackboard math symbols
\usepackage{amsmath}        % math environments
\usepackage{nicefrac}       % compact symbols for 1/2, etc.
\usepackage{microtype}      % microtypography
\usepackage{xcolor}         % colors
\usepackage{graphicx}       % graphics / resizebox
\usepackage{float}          % [H] float placement
\usepackage{multirow}       % table multirow
\usepackage{tikz}           % needed by pgfplots
\usepackage{pgfplots}
\usepackage{pgfplotstable}

\pgfplotsset{compat=newest}

\title{Efficient Modular Learning through Naive LoRA Summation: Leveraging Orthogonality in High-Dimensional Models}

\author{%
    Zhanhao Cao\thanks{Third year undergraduate student researcher in Professor Yingnian Wu's machine learning lab, under the guidance of PhD candidate Andrew Lizarraga} \\
    University of California, Los Angeles (UCLA) \\
    \texttt{richardcao@ucla.edu} \\
    \and
    Clement Truong\thanks{Second year undergraduate student researcher in Professor Yingnian Wu's lab} \\
    UCLA \\
    \texttt{clemtruong21@ucla.edu} \\
    \and
    Andrew Lizarraga\thanks{Fourth year PhD candidate} \\
    UCLA \\
    \texttt{andrewlizarraga@ucla.edu} \\
}

\begin{document}

\maketitle

\begin{abstract}
Recent advances in large language models (LLMs) are heavily supported by the size of their parameters, often reaching into the trillions. In parallel, parameter-efficient fine-tuning (PEFT) \cite{zhu2023surveypeft, meng2022mixreview} has emerged, allowing models to be fine-tuned by updating only a small fraction of parameters. One such method is Low-Rank Adaptation (LoRA) \cite{hu2021lora}, which, in the spirit of principal component analysis, updates only a small low-rank subspace. An important feature of LoRA is that it stores parameter deltas as the product of two small matrices rather than the full parameter dimension. This property enables treating LoRA deltas as efficient building blocks that can be appended to introduce domain expertise. Many prior works addressing interference when merging multiple domain-specific models \cite{lin2025lorasoups, chen2023loralego} have overlooked a property suggested by the superposition principle: independent matrices are likely to be nearly orthogonal \cite{frady2021superposition}. Leveraging this insight, we hypothesize that it is practical to train LoRA modules independently on disjoint domains, with their weight differences serving as building blocks that can be instantly appended to introduce specific knowledge through naive addition. Using GPT-2 Small (117M parameters) with LoRA rank 4 and $\alpha=64$, we fine-tuned adapters for three QA domains (math, medicine, finance). In pairwise tests, adding Math+Medicine adapters improved perplexity by $-9.10\%$ compared to merged-data fine-tuning, while Math+Finance and Finance+Medicine showed changes of $+4.54\%$ and $+27.56\%$, respectively. Furthermore, the RMS score between the LoRA deltas and the percent change in perplexity score shows a positive and linear relationship. This approach relies solely on simple addition, which can be completed in a few seconds, and achieves comparable performance to models trained on merged data.
\end{abstract}

\section{Introduction}
While LoRA was originally designed to reduce training and storage costs \cite{hu2021lora}, its structure implicitly supports modularity. Specifically, the low-dimension LoRA deltas are modules that can be appended to the base model. This perspective has invited many researchers to see if there are ways to merge multiple LoRA models to create multi-domain systems \cite{lin2025lorasoups, chen2023loralego}. However, most research on this topic is done using traditional, complex merging techniques. While those works have achieved desirable results, it is at the cost overhead in additional training or time-consuming computations. In contrast, this paper aims to have a naive approach to combine models by only adding the LoRA deltas. The hope is to use the linearity of LoRA updates to answer the question of whether addition could suffice in merging models of near orthogonal domains. If the superposition principle holds empirically for LoRA \cite{frady2021superposition, adler2024complexity}, it would allow for a simple and efficient strategy of training domain-specific LoRA modules in isolation, and use them as building blocks via direct summation. This would lead to a result of multi-domain adaptation without complex merging logic, alignment training, or even accessing the original dataset. This inquiry not only tests the practical feasibility of adapter composition but also sheds light on the structure of learned representations within LoRA. This paper would train multiple models and create all the additive combinations of them to evaluate the orthogonality of different domains, as well as independently analyzing the LoRA deltas to explore its potential orthogonality.

\section{Related Work}
\subsection*{LoRA Soups: Merging LoRAs for Practical Skill Composition Tasks \cite{lin2025lorasoups}}
This work investigates modularity in LoRA by introducing methods to combine multiple skill-specific adapters for complex task composition. Rather than operating directly on LoRA deltas, the authors fine-tune a weighted combination of full models derived from individually trained LoRAs. The paper demonstrates that averaging LoRA-induced full model weights, followed by additional fine-tuning, can outperform traditional model merging techniques. However, the approach requires access to downstream data and introduces additional compute costs during adaptation.

\subsection*{Decouple and Orthogonalize: A Data-Free Framework for LoRA Merging \cite{zheng2024decouple}}
Zheng et al. propose a data-free approach to LoRA merging by explicitly addressing interference among modules. They introduce a method that decouples magnitude and direction in LoRA updates, followed by lightweight optimization steps to enforce mutual orthogonality between components. This ensures minimal overlap and conflict between the merged deltas. The approach outperforms naive addition and prior LoRA merging baselines, especially when the source domains are semantically dissimilar or contain conflicting gradients.

\subsection*{Other notable methods}
\begin{itemize}
    \item O-LoRA \cite{sun2023olora}, which enforces orthogonality at the subspace level to support continual learning without forgetting.
    \item Mixture of LoRA Experts (MoLE) \cite{yan2024mole}, which dynamically selects from multiple LoRA modules via a routing mechanism for efficient multi-task adaptation.
\end{itemize}

These works collectively demonstrate possibility of composing LoRA deltas and models. However, most rely on either learned merging strategies or auxiliary optimization procedures.

\section{Experimental Setup}
The experiments uses GPT-2 Small (117M parameters) as the base model. It was chosen for its lower computational cost, with the experiment only asks for relative performances. Additionally, a smaller model is more susceptible to interference when merging LoRA modules due to collision of knowledge domain in smaller dimensions, making it a better candidate for highlighting the strengths and weaknesses of the proposed research. For fine-tuning dataset, we use three domain-specific question-answering (Q\&A) dataset: mathematics Q\&A data from Amini et al. \cite{amini2019mathqa}, medical Q\&A from the medQuad, and financial Q\&A from Hugging Face dataset. All dataset is consistent in Q\&A format for consistency. It's worth noting that such similarity might introduce correlation that could lead to constructive interference when summing LoRA deltas together. Each of the dataset is split in approximately 80:10:10 for training, validation, and testing respectively. The validation data is used solely for tuning hyperparameter and reducing overfitting. Using the calculated hyperparameter, the LoRA weights introduced by the training data would serve as the building block to support the experiment as described. Lastly, the testing data would be used to compute perplexity score, which the main metric to compute model capability used in this paper. The same methodologies would also be applied to each combination of datasets for more detailed analysis.

The LoRA fine-tuning process is done through the PEFT library, fixing LoRA rank and LoRA alpha in the configuration to ensure consistent delta scale and prevent domain-specific bias when summing the models. Referring to suggested values\cite{hu2021lora}, we have selected rank and alpha to be 4 and 64 respectively. These choices of hyperparameter is within the range of typical value, while promotes minimal weight updates in each building block to further pressure the idea of using limited dimension to achieve model training. Additionally, all models are trained and evaluated with maximum sequence length of 64, to standardize input processing. Fixing these hyperparameter is necessary to support unbiased comparison between models. Variables like dropout, batch size, and number of epochs will remain as hyperparameter tuned through validation dataset. The hyperparameter are found through Bayesian grid search technique from the Optuna library \cite{akiba2019optuna}. The models' training process follows the standard approach, using default GPT-2 tokenizer and default PEFT training configuration, which includes AdamW optimizer and learning rate scheduling. There is one crucial configuration to the trainer, that is only training the \texttt{c\_attn} and \texttt{c\_proj} layers of the GPT-2 model. This is inspired by the "Attention is All You Need" paper \cite{attention_all_you_need}, focusing only on the attention layer and the MLP, and this is in line with common LoRA implementation that aims for efficiency. Overall, this setup will generate controlled LoRA modules that will later be used in analysis and be treated as independent building blocks for composition experiments.

\section{Methodology}
Applying the training procedure described above, we fine-tuned the GPT-2 model on all seven partitions of the three datasets: individual domains (math, medicine, finance), pairs of two, and the composition of all three. For the models that involve more than one domain, data is concatenated prior to training. From each model, we extract the LoRA deltas. For the LoRA deltas from individual domains, we refer to them as \textit{fundamental building blocks}. On the other hand, LoRA deltas from composed models are referred to as \textit{level-k building blocks}, where k is the number of domains that are merged in the dataset. The extraction is done through a script that iterates through each trained \texttt{PeftModel} via its \texttt{named\_parameters}, selecting all \texttt{"lora"} parameters. This captures all injected LoRA layers, specifically limited to \texttt{c\_attn} and \texttt{c\_proj} layers.

The result of such extraction yields 36 layers of \texttt{lora\_A} ($\mathbf{A}$) and \texttt{lora\_B} ($\mathbf{B}$). Each layer represents a LoRA deltas matrix of shape $(n \times m)$, where the shapes of $\mathbf{A}$ and $\mathbf{B}$ are $(r \times m)$ and $(n \times r)$ respectively, with $r$ being the LoRA rank. Table~\ref{tab:lora_multiline} shows three different samples of what LoRA A and LoRA B look like, noting their stored shapes and the fact that LoRA weights are stored in a transposed and decomposed format \cite{hu2021lora}.

\begin{table}[h]
\small
\centering
\begin{tabular}{|l|l|}
\hline
\textbf{Field} & \textbf{Sample Content} \\
\hline
\textbf{Layer Name} & \texttt{base\_model.model.transformer.h.11.attn.c\_attn.lora\_A.default.weight} \\
\textbf{Shape} & $(4, 768)$ \\
\textbf{First 5 Values} & $[0.0048,\ -0.0167,\ 0.0321,\ 0.0039,\ 0.0242]$ \\
\hline
\textbf{Layer Name} & \texttt{base\_model.model.transformer.h.11.attn.c\_attn.lora\_B.default.weight} \\
\textbf{Shape} & $(2304, 4)$ \\
\textbf{First 5 Values} & $[-0.0024,\ -0.0023,\ -0.0036,\ -0.0014,\ -0.0078]$ \\
\hline
\textbf{Layer Name} & \texttt{base\_model.model.transformer.h.11.mlp.c\_proj.lora\_B.default.weight} \\
\textbf{Shape} & $(768, 4)$ \\
\textbf{First 5 Values} & $[-0.0049,\ 0.0053,\ -0.0049,\ 0.0045,\ 0.0003]$ \\
\hline
\end{tabular}
\caption{Three sample LoRA layers with shapes and first 5 values. Each block shows a distinct parameter group.}
\label{tab:lora_multiline}
\end{table}

To reconstruct the full LoRA delta matrix $\Delta \mathbf{W}$, we can compute the $n \times m$ matrix by applying the following equation:
\begin{equation}
\Delta \mathbf{W} = \left( \frac{\alpha}{r} \cdot \mathbf{B} \mathbf{A} \right)^\top
\end{equation}

Here, $\alpha$ is the LoRA scaling factor. This matrix represents the low-rank delta to be added to the frozen base model weights. These extracted deltas serve as the building blocks for our composition experiments. In this paper, we apply these building blocks by directly adding them to the original layer weights of the base model.

To better understand the application potential of the superposition principle, we can rewrite $\Delta \mathbf{W}$ as:
\begin{equation}
\Delta \mathbf{W} 
= \left( \frac{\alpha}{r} \cdot \mathbf{B} \mathbf{A} \right)^\top 
= \frac{\alpha}{r} \cdot \left( \sum_{i=1}^{r} \mathbf{A}[:, i] \cdot \mathbf{B}[i, :] \right)^\top
\end{equation}

This expression shows that the weight update is a sum of $r$ outer products, each of which is rank-1. This structure makes it especially relevant to the superposition principle, which suggests that a small number of low-rank, randomly oriented matrices are likely to remain approximately orthogonal in high-dimensional space \cite{frady2021superposition, cheung2019superposition}.

Here, $\mathbf{A}[:, i]$ is the $i$-th column of the matrix $\mathbf{A}$ (shape $r \times m$), and $\mathbf{B}[i, :]$ is the $i$-th row of matrix $\mathbf{B}$ (shape $n \times r$). For each fixed value of $i$, we see that each row is linearly dependent on one another, and each column is linearly dependent on another, resulting in a rank-1 matrix for a fixed $i$. Knowing this, we can discuss the rank of $\Delta \mathbf{W}$:
\[
\text{rank}(\Delta \mathbf{W}) 
\leq \sum_{i=1}^{r} \text{rank}(\mathbf{B}[i, :]^\top \cdot \mathbf{A}[:, i]^\top) 
= \sum_{i=1}^{r} 1 = r
\]

The above equation calculating the rank takes advantage of the fact that the rank of a sum of matrices is at most the sum of their individual ranks:
\[
\text{rank}\left(\sum_{i=1}^{r} \mathbf{M}_i\right) \leq \sum_{i=1}^{r} \text{rank}(\mathbf{M}_i)
\]

In our case, each matrix $\mathbf{B}[i, :]^\top \cdot \mathbf{A}[:, i]^\top$ is an outer product of two vectors and has rank 1. Therefore, summing $r$ such matrices results in a matrix of rank at most $r$ because of the following property from the Dimension Theorem of subspaces:
\[
\dim(A + B) = \dim(A) + \dim(B) - \dim(A \cap B) \quad \Rightarrow \quad
\dim(A + B) \leq \dim(A) + \dim(B)
\]

This result of an upper bound of $r$ is necessary for the application of the Superposition Principle, since the LoRA delta that we are using as a building block has dimension $r$, while the parameter space is $n \times m$. We have that $r \ll n \times m$, and the Superposition Principle should apply here. The experiment runs on the hypothesis that each building block is near orthogonal to each other. In this experiment, we will take the cosine similarity of each layer for the update, to see that the cosine similarity score is close to 0. More formally, we have each fundamental building block belong to the set $\mathcal{C}_r$, defined as follows:
\[
\mathcal{C}_r = \left\{ \mathbf{C} \in \mathbb{R}^{n \times m} \,\middle|\, \text{rank}(\mathbf{C}) \leq r \right\}
\]

Note that $\mathcal{C}_r$ is not a linear subspace. This is important because our experiment involves adding building blocks together, and the operation would not be closed under addition. Nevertheless, by applying the Dimension Theorem of subspaces again, the sum of two building blocks in $\mathcal{C}_r$ would be an element of $\mathcal{C}_{2r}$. In other words, by summing two data blocks together, the rank is at most $2r$. Similarly, adding $j$ different building blocks would create a rank of at most $\min(jr, n \times m)$. This shows that there could be an upper limit to how many building blocks could be applied before the model fails to maintain all the information simultaneously through this approach. The good news is that since $r \ll n \times m$, most value of $j$ would not cause any problems in terms of theoretical orthogonality, given that each matrix is independent from one another. With this in mind, our experiment combines all possible combinations of building blocks for analysis. These combinations are once again evaluated using cosine similarity to observe whether the similarity score increases as more building blocks are added. Specifically, we focus on level-k building blocks from two datasets, which would belong to the following subspace:
\[
\mathcal{C}_{2r} = \left\{ \mathbf{C} \in \mathbb{R}^{n \times m} \,\middle|\, \text{rank}(\mathbf{C}) \leq 2r \right\}
\]

The process above describes the methodology we use to construct the building blocks. These blocks will be combined and tested against the trivial building block retrieved directly from training on merged data. In addition, we will test each mixed-skill model on individual skills. Each combination of building blocks will also be evaluated with cosine similarity, measuring how the root mean squared of the cosine similarities across all 36 layers affects model performance. The single scalar we use to assess how much interference may exist between building blocks is given by:
\[
\text{RMS} = \sqrt{ \frac{1}{N} \sum_{i=1}^{N} \cos^2(\theta_i) }
\]

Finally, we conduct extensive tests on all 3 combined datasets using every building block combination and analyze the relationship between the RMS score and the model's perplexity.

\section{Result and Analysis}
First, we explore models built using fundamental building blocks versus those trained on merged data (control group) in two-domain tasks. Consider the Math-Medicine model. When using LoRA weights obtained from training on the merged dataset, the model achieves an average loss of 6.1798 and a perplexity of 482.8730. In contrast, applying both math and medicine building blocks separately to the base model yields an average loss of 6.0844 and a perplexity of 438.9414. This reflects a $-9.10\%$ change in perplexity when training with the addition of separately trained LoRA deltas. This result supports the feasibility of constructing multi-domain models via simple additive composition of domain-specific LoRA deltas. To assess the orthogonality between the math and medicine building blocks, we compute the cosine similarity between each of their LoRA deltas, including every layer in the Table~\ref{tab:cosine_similarity_math_medicine}.

\begin{table}[h!]
\centering
\small
\begin{tabular}{|c|c|c|c|}
\hline
\textbf{Layer} & \texttt{attn.c\_attn} & \texttt{attn.c\_proj} & \texttt{mlp.c\_proj} \\
\hline
0  & 0.007027 & 0.012169 & 0.055823 \\
1  & 0.009797 & 0.016873 & 0.045892 \\
2  & 0.001872 & 0.013048 & 0.044181 \\
3  & 0.003184 & 0.015216 & 0.092307 \\
4  & 0.004927 & 0.022680 & 0.106178 \\
5  & 0.007203 & 0.010741 & 0.073306 \\
6  & -0.000186 & -0.000942 & 0.106960 \\
7  & 0.008358 & 0.004753 & 0.072161 \\
8  & 0.005222 & 0.008017 & 0.093688 \\
9  & 0.015761 & 0.004821 & 0.128207 \\
10 & 0.005908 & 0.022097 & 0.121735 \\
11 & 0.013195 & 0.023702 & 0.050785 \\
\hline
\end{tabular}
\caption{Cosine similarities for LoRA deltas by layer for math and medicine}
\label{tab:cosine_similarity_math_medicine}
\end{table}

Analyzing these scores, we observe that most layers show positive cosine similarity, indicating that updates from both domains are generally in the same direction. Notably, the \texttt{mlp.c\_proj} layers exhibit higher peaks in similarity, suggesting overlap in the transformations applied after attention is computed. This is consistent with the shared structure of both datasets, as they are both framed in a question-answering (Q\&A) format and that MLP layers are responsible for integrating feature-level information. However, there can be other reasons that this experiment cannot explore. To further quantify the similarity, the RMS score between math and medicine building blocks is 0.0514. 

With the math/medicine combination serving as an example, we apply this computation across other two-domain combinations, shown in Table~\ref{tab:merged_vs_summed}.

\begin{table}[H]
\centering
\resizebox{\textwidth}{!}{%
\begin{tabular}{|l|c|c|c|c|}
\hline
\textbf{Combination} & \textbf{Perplexity (Merged)} & \textbf{Perplexity (Summed)} & \textbf{RMS Cosine Similarity} & \textbf{\% Change in Perplexity} \\
\hline
Math + Medicine      & 482.87 & 438.94 & 0.0514 & -9.10\% \\
Math + Finance       & 93.83  & 98.09  & 0.0583 & +4.54\% \\
Finance + Medicine   & 205.62 & 262.29 & 0.0708 & +27.56\% \\
\hline
\end{tabular}
}
\caption{Perplexity score of merged training vs model composed with fundamental building blocks}
\label{tab:merged_vs_summed}
\end{table}

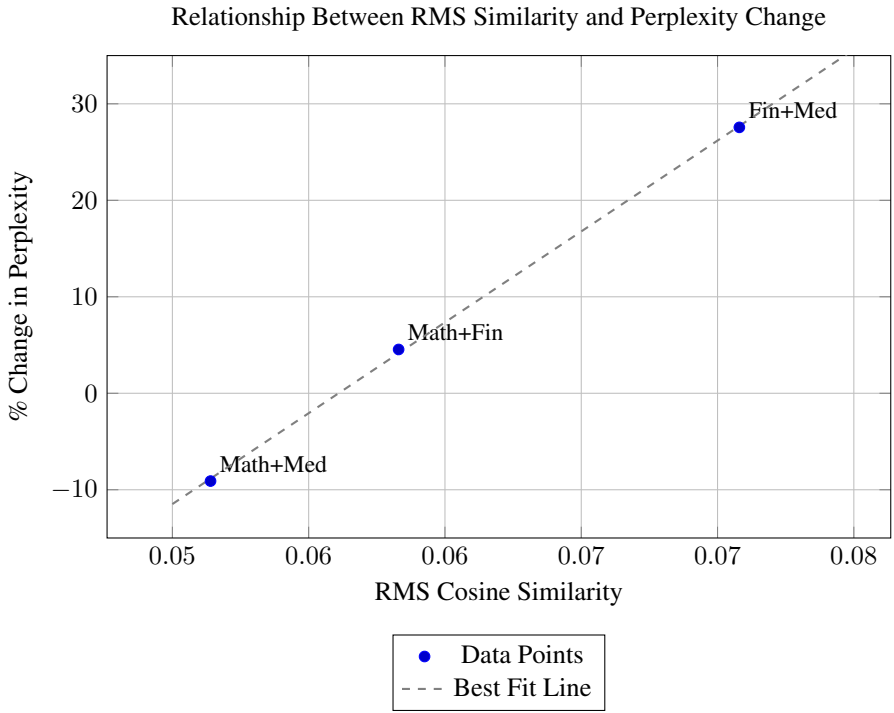
\begin{figure}[h!]
\centering
\begin{tikzpicture}
\begin{axis}[
    width=12cm,
    height=8cm,
    xlabel={RMS Cosine Similarity},
    ylabel={\% Change in Perplexity},
    title={Relationship Between RMS Similarity and Perplexity Change},
    grid=major,
    scaled x ticks=false,
    xticklabel style={/pgf/number format/fixed},
    legend style={at={(0.5,-0.2)},anchor=north},
    ymin=-15, ymax=35
]
% Data points
\addplot+[only marks, mark=*, color=blue] coordinates {
    (0.0514, -9.10)
    (0.0583, 4.54)
    (0.0708, 27.56)
};
\addlegendentry{Data Points}
% Annotations
\node at (axis cs:0.0514,-9.10) [anchor=south west]{\small Math+Med};
\node at (axis cs:0.0583,4.54) [anchor=south west]{\small Math+Fin};
\node at (axis cs:0.0708,27.56) [anchor=south west]{\small Fin+Med};
\addplot[domain=0.05:0.075, dashed, thick, gray] {1883.89*x - 105.68};
\addlegendentry{Best Fit Line}
\end{axis}
\end{tikzpicture}
\caption{RMS similarity vs. \% change in perplexity for pairwise domain combinations.}
\label{fig:linear_fit_rms_vs_perplexity}
\end{figure}

From these three combinations of two-domain models, we observe limited/moderate perplexity changes in all cases. This suggests the feasibility of summing LoRA deltas directly as an efficient method to combine capabilities across domains without requiring additional retraining. Moreover, we observe that the RMS cosine similarity remains relatively low for all three combinations. This supports the assumption of near-orthogonality in LoRA deltas, as theorized by the Superposition Principle. Notably, there appears to be a positive relationship between RMS cosine similarity and the percentage change in perplexity. This trend aligns with theoretical expectations: higher RMS scores (ranging from 0 to 1) suggest that the delta matrices are less orthogonal, increasing the likelihood of destructive interference during composition. While these results are based on only three data points and cannot be fully generalized, the data points appear to be nearly collinear. This suggests a potential linear relationship, which may become clearer with additional data. The best-fitting line for these three points is shown in Figure~\ref{fig:linear_fit_rms_vs_perplexity}.

\[
y = 1883.89x - 105.68
\]

We now explore building 3-domain models through additive building blocks. In this case, the baseline model is trained on the merged data of all three datasets, achieving a perplexity of \textbf{127.76}. There are four different ways to construct 3-domain models through composition. Three strategies involve adding a level-2 building block with the remainder fundamental block, and another through adding three fundamental building blocks. Here are the results of each strategy:

\begin{table}[h]
\centering
\resizebox{\textwidth}{!}{%
\begin{tabular}{|l|c|c|c|c|}
\hline
\textbf{Combination} & \textbf{Perplexity (Merged)} & \textbf{Perplexity (Summed)} & \textbf{RMS Cosine Similarity} & \textbf{\% Change in Perplexity} \\
\hline
Med + Fin + Math     & 127.76 & 191.28 & --     & +49.67\% \\
Other 2 + Math       & 127.76 & 238.52 & 0.0714 & +86.60\% \\
Other 2 + Med        & 127.76 & 194.12 & 0.0824 & +51.94\% \\
Other 2 + Fin        & 127.76 & 260.35 & 0.0872 & +103.76\% \\
\hline
\end{tabular}
}
\caption{Perplexity comparisons for 3-domain additive models}
\label{tab:3domain_comparison}
\end{table}

Summing all three LoRA deltas results in a perplexity of 191.28, which is the best performance out of the four methods of applying building blocks. However, there's a +49.67\% perplexity score increase compared to merged training, much higher than the 2-domain cases. This is likely due to the result of constructive interference in limited parameter space. Nevertheless, the score is still comparable to the control model trained on merged data. Meanwhile, it's important to not neglect the undesired effect of constructive interference in similar tasks, which can be eliminated when using more distinct datasets or utilizing more parameter space for training. The set up of the experiment begs for such interference through using the same question format, relatively small base model, and minimal training parameter dimension.

The performance from level-2 building block with fundamental block isn't great, with the worst-performing composition being (Medicine + Math) + Finance, corresponding to a +103.76\% increase from the control model. Additionally, there no longer appears to be a clear linear relationship with RMS similarity as in the 2-domain cases. This is likely because the level-2 building block  compresses two domains into a low-rank approximation, while the fundamental building block can use a full rank, which is more emphasized than the other two domains. Such result highlight a key limitation of level-k building block of not treating such domain equally. To make level-k building block more useful, more advanced merging techniques must be used.

These findings suggest that while LoRA delta addition remains a viable strategy, it becomes less effective as more domains are added. Storing multiple domain-specific deltas in a limited-rank representation might introduces interference, thus independent dataset is the key to make naive addition work. This is largely because superposition principle assumes approximate orthogonality among low-rank updates, which becomes harder to maintain with overlapping domain tasks \cite{frady2021superposition, cheung2019superposition}.

Lastly, we examine the cosine similarity details across layers. Once again, we observe spikes in the \texttt{mlp.c\_proj} layers. In contrast, the \texttt{attn.c\_attn} and \texttt{attn.c\_proj} layers conform more closely to the expectations of the superposition principle, suggesting that attention submodules can independently learn different domain-specific behaviors. Such independence may even allow additive updates to outperform traditional merged training, as was seen in the Math + Medicine case. However, true task separation—especially in downstream MLP transformations—appears to be a limiting factor in this experiment.

\begin{table}[H]
\centering
\small
\resizebox{\textwidth}{!}{%
\begin{tabular}{|c|ccc|ccc|ccc|}
\hline
\multirow{2}{*}{\textbf{Layer}} 
& \multicolumn{3}{c|}{\textbf{2+Fin}} 
& \multicolumn{3}{c|}{\textbf{2+Med}} 
& \multicolumn{3}{c|}{\textbf{2+Math}} \\
& \texttt{attn.c\_attn} & \texttt{attn.c\_proj} & \texttt{mlp.c\_proj} 
& \texttt{attn.c\_attn} & \texttt{attn.c\_proj} & \texttt{mlp.c\_proj} 
& \texttt{attn.c\_attn} & \texttt{attn.c\_proj} & \texttt{mlp.c\_proj} \\
\hline
0  & 0.0034 & 0.0125 & 0.1337 & 0.0068 & 0.0207 & 0.1182 & 0.0063 & 0.0087 & 0.0858 \\
1  & 0.0192 & 0.0121 & 0.0905 & 0.0203 & 0.0209 & 0.0723 & 0.0121 & 0.0150 & 0.0787 \\
2  & 0.0266 & 0.0195 & 0.0710 & 0.0158 & 0.0172 & 0.0606 & 0.0124 & 0.0212 & 0.0697 \\
3  & 0.0108 & 0.0237 & 0.1607 & 0.0034 & 0.0213 & 0.1397 & 0.0123 & 0.0237 & 0.1421 \\
4  & 0.0190 & 0.0350 & 0.1778 & 0.0186 & 0.0353 & 0.1627 & 0.0052 & 0.0314 & 0.1551 \\
5  & 0.0127 & 0.0244 & 0.1672 & 0.0150 & 0.0173 & 0.1534 & 0.0068 & 0.0237 & 0.1103 \\
6  & 0.0160 & 0.0184 & 0.1807 & 0.0132 & 0.0087 & 0.1746 & 0.0010 & 0.0102 & 0.1479 \\
7  & 0.0168 & 0.0257 & 0.1479 & 0.0159 & 0.0213 & 0.1416 & 0.0126 & 0.0095 & 0.1014 \\
8  & 0.0230 & 0.0253 & 0.1545 & 0.0151 & 0.0199 & 0.1581 & 0.0143 & 0.0163 & 0.1197 \\
9  & 0.0185 & 0.0212 & 0.1709 & 0.0208 & 0.0225 & 0.1873 & 0.0201 & 0.0035 & 0.1534 \\
10 & 0.0066 & 0.0225 & 0.1689 & 0.0063 & 0.0307 & 0.1613 & 0.0091 & 0.0230 & 0.1643 \\
11 & 0.0240 & 0.0307 & 0.1034 & 0.0178 & 0.0369 & 0.0776 & 0.0250 & 0.0264 & 0.0723 \\
\hline
\end{tabular}
}
\caption{Layer-wise cosine similarities for three 2-domain LoRA combinations}
\label{tab:cosine_similarity_all_three}
\end{table}

\section{Future Improvement and Conclusion}
This work demonstrates that LoRA deltas themselves can be used as building blocks and combined additively to approximate multi-domain capabilities with minimal computation overhead. While performance degradation emerges as more building blocks are applied, the result is promising when the building blocks demonstrate orthogonal property. This property allows instant knowledge of a domain, which could be applied in models that require quick adaptation such as storing temporary memory. This additive property also has a simple inverse for unlearning, by the additive inverse property of the LoRA deltas subspace. Future work will explore scaling to larger models, where increased parameter space may enhance the orthogonality assumption. Additional investigations could include training on more diverse datasets to strengthen statistical trends. There direction aim to push LoRA summation from a feasible trick to a possibly principled paradigm for modular learning.

Future work will explore scaling to larger models, where increased parameter space may enhance the orthogonality assumption. Additional investigations could include training on more diverse datasets to strengthen statistical trends, as well as extending additive composition to other settings. For example, latent planning approaches such as the Latent Plan Transformer \cite{NEURIPS2024_df22a196} and Latent Adaptive Planner for Dynamic Manipulation \cite{noh2025latentadaptiveplannerdynamic} could inspire applications of modular LoRA composition in control and robotics. Similarly, techniques for efficient prompt compression \cite{honig2025betterpromptcompressionmultilayer} and temporal adaptation mechanisms such as Temp-LoRA, recently applied in SlowFast-VGen for long video generation \cite{hong2024slowfastvgenslowfastlearningactiondriven}, may provide pathways to improve scalability and adaptability in real-time generative systems. These directions aim to push LoRA summation from a feasible trick to a possibly principled paradigm for modular learning.


\begin{thebibliography}{10}

\bibitem{zhu2023surveypeft}
Ying Zhu, Changbo Gao, Hong Wu, Degen Huang, Yunlong Zhang, and Chuanxing Deng.
\newblock A comprehensive survey on parameter efficient transfer learning.
\newblock {\em arXiv preprint arXiv:2306.11644}, 2023.

\bibitem{meng2022mixreview}
Yanda Meng, Ze~Bai, Zhewei Yu, Weijia Xue, and Ji~Zhou.
\newblock Mixture-of-adapters: Efficiently fine-tuning language models for parameter-efficient transfer learning.
\newblock {\em arXiv preprint arXiv:2205.00049}, 2022.

\bibitem{hu2021lora}
Edward~J Hu, Yelong Shen, Phil Wallis, Zeyuan Allen-Zhu, Yuanzhi Li, Weizhu Wang, and Zhi Chen.
\newblock Lora: Low-rank adaptation of large language models.
\newblock {\em arXiv preprint arXiv:2106.09685}, 2021.

\bibitem{lin2025lorasoups}
Ying Lin, Yixuan Wang, Wei-Lun Chiang, Chitta Baral, and Xiaoqiang Xie.
\newblock Lora soups: Merging loras for practical skill composition tasks.
\newblock In {\em Proceedings of the 63rd Annual Meeting of the Association for Computational Linguistics: Industry Track (ACL Industry)}, 2025.

\bibitem{chen2023loralego}
Shiyang Chen, Yongliang Zhao, Weiguo Yang, Zhihao Tan, and Xiaoguang Yu.
\newblock Merging loras like playing lego: Pushing the modularity of lora to extremes through rank-wise clustering.
\newblock {\em arXiv preprint arXiv:2312.00466}, 2023.

\bibitem{frady2021superposition}
E.~Paxon Frady, Denis Kleyko, and Friedrich~T Sommer.
\newblock Theory of the superposition principle for randomized connectionist representations in neural networks.
\newblock {\em Nature Communications}, 12(1):1--14, 2021.

\bibitem{adler2024complexity}
Micah Adler and Nir Shavit.
\newblock On the complexity of neural computation in superposition.
\newblock {\em arXiv preprint arXiv:2409.15318}, 2024.

\bibitem{zheng2024decouple}
Shenghe Zheng, Hongzhi Wang, Chenyu Huang, Xiaohui Wang, Tao Chen, Jiayuan Fan, Shuyue Hu, and Peng Ye.
\newblock Decouple and orthogonalize: A data-free framework for lora merging.
\newblock {\em arXiv preprint arXiv:2505.15875}, 2024.

\bibitem{sun2023olora}
Zixuan Sun, Lingkai Kong, Zhiyuan Li, Jieyu Gao, Yiming Ma, Bill~Yuchen Lin, and Xiang Ren.
\newblock O-lora: Orthogonal subspace learning for language model continual learning.
\newblock {\em arXiv preprint arXiv:2310.14152}, 2023.

\bibitem{yan2024mole}
Bowen Yan, Yixiao Chen, Yikang Jiang, Haoyi Bai, Zeqi Lin, Xudong Li, Philip~S Yu, Yingyao Chen, and Zhewei Wang.
\newblock Mixture of lora experts.
\newblock {\em arXiv preprint arXiv:2404.13628}, 2024.

\bibitem{amini2019mathqa}
Ayyappan Amini, Saad Gabriel, Peter Lin, Rik Koncel-Kedziorski, and Hannaneh Hajishirzi.
\newblock Mathqa: Towards interpretable math word problem solving with operation-based formalisms.
\newblock {\em arXiv preprint arXiv:1905.13319}, 2019.

\bibitem{akiba2019optuna}
Takuya Akiba, Shotaro Sano, Takeru Yanase, Toshihiko Ohta, and Masanori Koyama.
\newblock Optuna: A next-generation hyperparameter optimization framework.
\newblock In {\em Proceedings of the 25th ACM SIGKDD International Conference on Knowledge Discovery \& Data Mining}, pages 2623--2631. ACM, 2019.

\bibitem{attention_all_you_need}
Ashish Vaswani, Noam Shazeer, Niki Parmar, Jakob Uszkoreit, Llion Jones, Aidan~N Gomez, {\L}ukasz Kaiser, and Illia Polosukhin.
\newblock Attention is all you need.
\newblock In I.~Guyon, U.~Von Luxburg, S.~Bengio, H.~Wallach, R.~Fergus, S.~Vishwanathan, and R.~Garnett, editors, {\em Advances in Neural Information Processing Systems}, volume~30. Curran Associates, Inc., 2017.

\bibitem{cheung2019superposition}
Brian Cheung, Alex Terekhov, Yubei Chen, Pulkit Agrawal, and Bruno~A Olshausen.
\newblock Superposition of many models into one.
\newblock {\em arXiv preprint arXiv:1902.05522}, 2019.

\bibitem{NEURIPS2024_df22a196}
Deqian Kong, Dehong Xu, Minglu Zhao, Bo~Pang, Jianwen Xie, Andrew Lizarraga, Yuhao Huang, Sirui Xie, and Ying~Nian Wu.
\newblock Latent plan transformer for trajectory abstraction: Planning as latent space inference.
\newblock In A.~Globerson, L.~Mackey, D.~Belgrave, A.~Fan, U.~Paquet, J.~Tomczak, and C.~Zhang, editors, {\em Advances in Neural Information Processing Systems}, volume~37, pages 123379--123401. Curran Associates, Inc., 2024.

\bibitem{noh2025latentadaptiveplannerdynamic}
Donghun Noh, Deqian Kong, Minglu Zhao, Andrew Lizarraga, Jianwen Xie, Ying~Nian Wu, and Dennis Hong.
\newblock Latent adaptive planner for dynamic manipulation, 2025.

\bibitem{honig2025betterpromptcompressionmultilayer}
Edouardo Honig, Andrew Lizarraga, Zijun~Frank Zhang, and Ying~Nian Wu.
\newblock Better prompt compression without multi-layer perceptrons, 2025.

\bibitem{hong2024slowfastvgenslowfastlearningactiondriven}
Yining Hong, Beide Liu, Maxine Wu, Yuanhao Zhai, Kai-Wei Chang, Linjie Li, Kevin Lin, Chung-Ching Lin, Jianfeng Wang, Zhengyuan Yang, Yingnian Wu, and Lijuan Wang.
\newblock Slowfast-vgen: Slow-fast learning for action-driven long video generation, 2024.

\end{thebibliography}
\end{document}